%% file: main.tex
\documentclass[]{fairmeta}
\usepackage{pifont}
\usepackage{amsmath}
\usepackage{amssymb}
\usepackage{makecell} 
\usepackage{atbegshi}
\usepackage{mathtools}
\usepackage{amsthm}
\usepackage{enumitem}
\usepackage{algorithm}
\usepackage{algpseudocode}
\usepackage{booktabs}
\usepackage{array}
\usepackage{booktabs}
\usepackage{array}
\usepackage{makecell}

\usepackage{tabularx}
\usepackage{makecell}
\usepackage{booktabs}
\usepackage[most]{tcolorbox}
\usepackage{xcolor}
\usepackage{fvextra}
\usepackage{xurl}
\usepackage{hyperref}
\usepackage{fontspec}
\usepackage{eso-pic}

\newcolumntype{L}[1]{>{\raggedright\arraybackslash}m{#1}}
\newcolumntype{C}[1]{>{\centering\arraybackslash}m{#1}}

\newtcolorbox{templatebox}[1]{
    breakable,
    enhanced,
    colback=white,
    colframe=gray!80!black,
    colbacktitle=gray!80!black,
    coltitle=white,
    fonttitle=\bfseries,
    title=#1,
    arc=3mm,
    boxrule=1pt,
    drop fuzzy shadow={gray!50!white},
    left=5mm,
    right=5mm,
    top=3mm,
    bottom=3mm
}

\renewcommand{\affiliation}[2][]{%
  \addtolist[#1]{#2}{\affiliationlist}{\affiliationformat}{\quad}%
}

\title{%
SkillClaw: Let Skills Evolve Collectively with Agentic Evolver
}

\author[*]{Ziyu Ma}
\author[*]{Shidong Yang}
\author[*]{Yuxiang Ji}
\author[*]{Xucong Wang}
\author[\dagger]{Yong Wang}
\author[]{Yiming Hu}
\author[]{Tongwen Huang}
\author[]{Xiangxiang Chu}

\affiliation[ ]{AMAP, Alibaba Group}
\affiliation[*]{Equal contribution}
\affiliation[\dagger]{Project lead}

\abstract{%
Large language model (LLM) agents such as OpenClaw rely on reusable skills to perform complex tasks, yet these skills remain largely static after deployment. As a result, similar workflows, tool usage patterns, and failure modes are repeatedly rediscovered across users, preventing the system from improving with experience. While interactions from different users provide complementary signals about when a skill works or fails, existing systems lack a mechanism to convert such heterogeneous experiences into reliable skill updates. To address these issues, we present SkillClaw, a framework for collective skill evolution in multi-user agent ecosystems, which treats cross-user and over-time interactions as the primary signal for improving skills. SkillClaw continuously aggregates trajectories generated during use and processes them with an autonomous evolver, which identifies recurring behavioral patterns and translates them into updates to the skill set by refining existing skills or extending them with new capabilities. The resulting skills are maintained in a shared repository and synchronized across users, allowing improvements discovered in one context to propagate system-wide while requiring no additional effort from users. By integrating multi-user experience into ongoing skill updates, SkillClaw enables cross-user knowledge transfer and cumulative capability improvement, and experiments on WildClawBench show that limited interaction and feedback, it significantly improves the performance of Qwen3-Max in real-world agent scenarios.
}

\metadata[Github]{\url{https://github.com/AMAP-ML/SkillClaw}}

\begin{document}
\maketitle

\section{Introduction}
\label{sec:intro}

Large language model (LLM) agents~\citep{yao2022react,shinn2023reflexion} have rapidly made personal AI assistants practical in real-world settings, with systems such as OpenClaw enabling users to complete complex tasks through natural conversation. A user can now ask an agent to configure a service, debug an API call, or automate a multi-step workflow, relying on it to coordinate tool usage and intermediate reasoning. These capabilities are largely driven by skills, which encode structured procedures for interacting with tools and solving tasks. In current deployments, users typically select and install skills from a centralized skill hub to meet their needs, and these skills serve as the primary building blocks for agent behavior. However, the skill ecosystem remains largely static~\citep{zhang2025agentracer,naihin2023testing,song2026agents}, as skills are manually installed and maintained and solutions discovered during interaction rarely persist beyond individual sessions. 

This limitation becomes evident in everyday usage. For example, users often ask agents to complete multi-step tasks such as automating data processing workflows, where failures frequently arise from subtle issues such as incorrect argument formats or mismatched tool calls. Through several rounds of trial and error, an agent may eventually arrive at a working solution or even a more stable procedure. However, these improvements remain confined to the current session and are not consolidated into the skill set or carried forward to future interactions. As similar tasks recur across different users and over time, the same patterns of failure and recovery are repeatedly observed, yet the system does not improve its behavior. This is fundamentally problematic because users operate in overlapping task spaces where similar workflows, tools, and failure modes are shared, but the system fails to leverage these recurring experiences. Consequently, each user is forced to rediscover solutions independently, preventing knowledge from accumulating at the system level. Therefore, the key challenge is not only to improve performance within a single session, but also to enable knowledge to accumulate and evolve across users.

Existing approaches to agent adaptation fail to support the accumulation and evolution of skills across users and over time. Memory-based methods store past trajectories for retrieval~\citep{shinn2023reflexion,zhao2024expel,fang2025memp,tang2025agent,ouyang2025reasoningbank,chhikara2025mem0,liu2026simplemem}, but such records remain tied to specific instances and are difficult to generalize into improved behavior. Skill-based methods compress experience into structured instructions~\citep{xia2026skillrl,zhang2025memevolve,zhang2026memrl,wu2025evolver,zhang2026memskill}, yet treat the resulting skill library as a static resource that does not evolve through usage. While local refinement can improve individual agent instances, these improvements remain isolated and do not accumulate across users, leading to fragmented skills rather than collective improvement over time. What is missing is a mechanism that turns ordinary interactions into continuous skill evolution and enables skills to improve collectively across users.

\begin{figure*}[t] 
    \centering
    \includegraphics[width=1\linewidth]{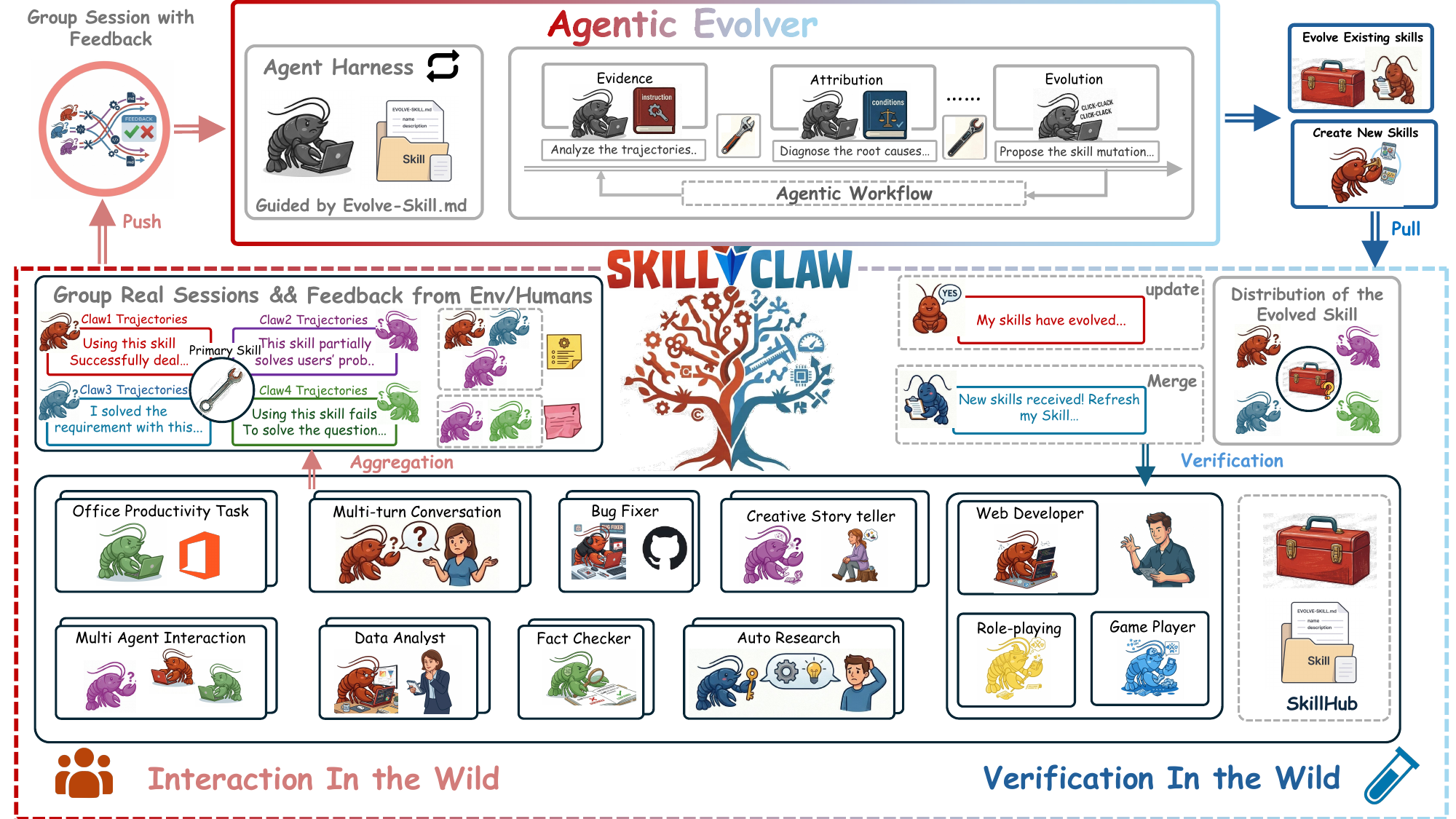} 
    \vspace{-0.6cm}
    \caption{\textbf{Overview of SkillClaw. }
SkillClaw enables collective skill evolution in a multi-user agent ecosystem through a closed-loop pipeline. Independent agents interact with their environments and produce structured session trajectories that preserve full action–feedback causal chains. These trajectories are aggregated across users and grouped by referenced skills, forming a shared evidence base that exposes consistent success patterns and recurring failure modes. An agentic evolver analyzes each skill-specific group and performs evidence-driven updates via refinement or creation, while preserving validated behaviors from successful executions. The updated skill repository is then synchronized back to all agents, allowing improvements discovered in one user’s interaction to benefit others and continuously accumulate over time.} 
    \vspace{-0.2cm}
    \label{fig:main}
\end{figure*}  

Building on this insight, we propose \textbf{SkillClaw}, a framework for skill collective evolution in multi-user OpenClaw-style agent ecosystems (Fig ~\ref{fig:main}). SkillClaw adopts a centralized evolution architecture, where agents deployed across different users continuously generate interaction sessions during everyday usage. These trajectories are aggregated across users and over time as evidence of real-world task execution and are processed by a centralized evolution engine to drive skill updates. Given accumulated interaction trajectories, the evolver analyzes both successful and failed executions, identifies recurring issues and effective procedures, and updates the shared skill set by refining existing skills, creating new ones, or adjusting their descriptions. Unlike predefined pipelines, this evolution process is driven by an autonomous agent that performs open-ended reasoning over interaction evidence and directly edits skill definitions. The updated skills are then synchronized across agents, allowing improvements discovered in one context to propagate to future interactions across users and over time. This forms a continuous evolution loop in which interaction data drives skill updates, and updated skills improve subsequent interactions. From the user’s perspective, this process requires no additional effort, as data collection, evolution, and synchronization all occur automatically in the background.

This design introduces three key properties that distinguish SkillClaw from existing systems. First, SkillClaw enables \emph{collective evolution}, where knowledge from individual interactions contributes to a shared and continuously improving skill ecosystem. Second, it is \emph{fully automatic}, with skill evolution driven by runtime interaction without manual curation or explicit user intervention. Third, it adopts an \emph{agentic evolution paradigm}, where skill updates are produced through open-ended reasoning rather than predefined update rules, enabling flexible and context-aware improvements.

SkillClaw is designed as a general framework that is compatible with a wide range of Claw-style agent systems, including OpenClaw as well as variants such as CoPaw, IronClaw, PicoClaw, ZeroClaw, NanoClaw, and NemoClaw. We evaluate SkillClaw on WildClawBench using \texttt{qwen3-max} as the backbone model and simulate a multi-user deployment setting. 
Experimental results demonstrate that SkillClaw yields substantial improvements across tasks, highlighting the effectiveness of multi-user driven collective evolution for building continuously improving agent systems in real-world environments.

\input{sec/method}

\input{sec/experiment}
\input{sec/related_work}

\bibliographystyle{plainnat}
\bibliography{ref}

\input{sec/appendix}

\end{document}

%% file: sec/method.tex
\section{Method}
\label{sec:method}

We present \textbf{SkillClaw}, a framework for collective skill evolution in a multi-user agent ecosystem (Fig ~\ref{fig:main}). 
In our setting, different users independently interact with their own deployed OpenClaw agents, potentially across different devices, environments, and time. 
Although these interactions are isolated at runtime, they share a common behavioral space: similar workflows, overlapping tool usage, and recurring failure modes appear across users.
SkillClaw builds on the observation that \emph{different users exercising the same skill under diverse contexts produce complementary views of that skill's behavioral boundary}, revealing both the conditions under which it works and those under which it breaks.
A single user rarely generates enough signal to separate a generalizable improvement from an idiosyncratic fix. Aggregating evidence across users provides the grounding that makes stable skill evolution possible.

Formally, let $\mathcal{S} = \{s_1, \dots, s_M\}$ denote a shared skill set, where each skill is a reusable procedural artifact. 
Each user interaction produces a session trajectory $\tau$, which records the full interaction loop: the prompt, the agent's actions, feedback from the environment or the user, and the final agent response.
Given a set of trajectories $\mathcal{T} = \{\tau_i\}$ collected across users, our goal is to update the shared skill set:
\[
\mathcal{S}' = \Phi(\mathcal{S}, \mathcal{T}),
\]
such that improvements discovered in one interaction can benefit future users.

\subsection{From Isolated Sessions to Shared Evidence}
\label{sec:evidence}

Multi-user skill evolution requires converting a stream of isolated, heterogeneous interaction sessions into a form that supports cross-user reasoning.
SkillClaw does this in two stages: it first structures individual sessions to preserve causal information, then aggregates them into a shared evidence base.

At the system level, SkillClaw connects independently deployed agents through a common skill repository.
Each agent has access to the current skill set and produces interaction sessions during normal usage. These sessions are recorded and uploaded as shared evidence.
A centralized evolution engine periodically processes the collected sessions, updates the skill repository, and synchronizes the updated skills back to all agents, forming a closed loop:
\[
\text{Multi-user Interaction} \rightarrow 
\text{Session Collection} \rightarrow 
\text{Skill Evolution} \rightarrow 
\text{Skill Synchronization}.
\]
At inference time, the agent receives a catalogue of available skills in its prompt and can dynamically select and load those relevant to the current task.
Users do not interact directly, and no coordination among agents is required. Collective improvement arises entirely from shared skill evolution.

Within this loop, each interaction session contains more than plain dialogue.
SkillClaw records the full causal chain: the user prompt, the agent's actions (including tool calls), intermediate feedback (tool results, errors, and explicit user responses), and the final agent response.
We record all of this because most skill-level failures are \emph{procedural}. An incorrect argument format, a missing validation step, or a misordered tool call can cause a task to fail, yet none of these problems appears in the final response. They can only be diagnosed from the intermediate action-feedback trace.
Each raw session is converted into a structured representation that preserves this chain:
\[
\text{prompt} \rightarrow \text{action} \rightarrow \text{feedback} \rightarrow \cdots \rightarrow \text{agent response}.
\]
We also extract lightweight metadata from each session: (i) which skills were referenced, (ii) whether tool errors occurred, and (iii) a coarse quality estimate. 
These signals help organize sessions but do not impose rigid labels.

Once sessions are structured, they are grouped by the skills they reference to enable cross-user reasoning. 
For each skill $s$, we collect all sessions that invoked $s$:
\[
\mathcal{G}(s) = \{ \tau_i \mid s \in \mathcal{K}_i \},
\]
and place sessions that did not use any skill into a separate group $\mathcal{G}(\varnothing)$.
This grouping does more than organize the data.
When multiple sessions invoke the same skill but produce different outcomes across different users, tasks, or environments, the comparison directly reveals where the skill works and where it fails, with the skill itself as the controlled factor.
This amounts to a \emph{natural ablation} and enables two operations that would be unreliable from single-user data alone: 
(1) evaluating how an existing skill actually performs under diverse real-world usage, and 
(2) identifying recurring procedures that no existing skill covers, surfaced by patterns in $\mathcal{G}(\varnothing)$.

\begin{algorithm}[t]
\caption{Agentic Collective Skill Evolution}
\label{alg:agentic_evolution}
\begin{algorithmic}[1]
\Require Skill repository $\mathcal{S}$, user sessions $\mathcal{T}$
\Ensure Updated repository $\mathcal{S}'$

\State Convert $\mathcal{T}$ into structured evidence $\mathcal{E}$
\State Group $\mathcal{E}$ by referenced skills to obtain $\{\mathcal{G}(s)\}$ and $\mathcal{G}(\varnothing)$
\State $\mathcal{S}' \gets \mathcal{S}$

\ForAll{group $\mathcal{G}(s)$}
    \State Use the agentic evolver to analyze recurring success and failure patterns
    \State Select an evolution action from \{\texttt{refine}, \texttt{create}, \texttt{skip}\}
    \State Generate a candidate skill update if the evidence supports modification
    \State Apply conservative editing and validation
    \State Merge approved updates into $\mathcal{S}'$
\EndFor

\State Analyze $\mathcal{G}(\varnothing)$ for missing but reusable procedures
\State Add validated new skills into $\mathcal{S}'$
\State Synchronize $\mathcal{S}'$ back to all agents
\State \Return $\mathcal{S}'$
\end{algorithmic}
\end{algorithm}

\subsection{Agentic Skill Evolution}
\label{sec:evolution}
The core of SkillClaw is an \emph{agentic evolver} that updates the shared skill repository with open-ended reasoning. SkillClaw instantiate an \emph{agentic evolver}, an LLM agent equipped with a structured harness that supplies the grouped session evidence, the current skill definitions, and a set of permitted evolution actions. The harness provides structured inputs but does not constrain the evolver's reasoning. The evolver diagnoses root causes from sessions of varying context lengths and skills of different formats, and decides how to act. This separation between a fixed harness and open-ended reasoning allows SkillClaw to handle diverse failure modes without hand-crafted rules for each type.

Concretely, given a skill $s$ and its associated session group $\mathcal{G}(s)$, the evolver examines both successful and failed executions and selects one of three actions:
\begin{itemize}
    \item \textbf{Refine.} Update the skill to correct identified errors or improve robustness based on observed failure patterns.
    \item \textbf{Create.} Introduce a new skill when $\mathcal{G}(s)$ reveals recurring sub-procedures that are not captured by any existing skill.
    \item \textbf{Skip.} Leave the skill unchanged when the available evidence is insufficient to justify a modification.
\end{itemize}
For sessions in $\mathcal{G}(\varnothing)$, i.e., those that did not invoke any skill, the evolver focuses on discovering missing but reusable procedures. 
New skills are created only when the observed patterns are specific enough to be teachable and likely to recur.

Regardless of which action is chosen, the evolver always reasons over successful and failed sessions \emph{jointly}. Successful sessions define the \emph{invariants} of a skill, the parts that work and must not be altered. Failed sessions define the \emph{targets}, the specific behaviors that need correction.
This joint view is what prevents a naive failure: fixing one problem while inadvertently breaking a previously effective procedure. Each update corrects identified deficiencies while preserving what successful sessions have validated, making evolution cumulative.
The complete procedure is given in Algorithm~\ref{alg:agentic_evolution}.




\subsection{Skill Synchronization and the Evolution Loop}
\label{sec:loop}

After evolution, candidate skill updates are validated before being written back to the shared repository. Validation is performed during the nighttime and executed in available idle user environments, ensuring that evaluation reflects real deployment conditions. For a skill $s$ and its candidate update $s'$, the system selects relevant tasks from the interaction data collected during the day. Both versions are executed under the same environment using the full toolchain, including multi-step interactions and intermediate feedback. After execution, the system uses the model to compare the outcomes produced by $s$ and $s'$. The decision is based on overall task success and execution stability. If the updated skill demonstrates better performance, it is marked as \texttt{Accept}; otherwise, it is marked as \texttt{Reject}. Accepted updates are merged into the shared repository and synchronized to all agents for the next day. Rejected updates are retained only as candidates and are not deployed. As a result, users always interact with the best validated skill pool from the previous night, rather than unverified updates. This validation step induces a monotonic deployment behavior. Since only improvements are accepted, the deployed skill pool does not degrade over time. Combined with the evolution process, the system forms a closed loop:
\[
\text{Interaction} \rightarrow \text{Evidence} \rightarrow \text{Evolution} \rightarrow \text{Validation} \rightarrow \text{Deployment}.
\]

where updated skills shape future interactions and generate new evidence for the next round of evolution.

Three properties follow from this design. First, \emph{collective evolution}. Sessions are aggregated across users, and knowledge discovered in one interaction is propagated to a shared skill ecosystem that benefits all users.
Second, \emph{full automation}. The entire pipeline, from session recording to skill synchronization, runs without manual curation or explicit user intervention. The only human input is normal agent usage.
Third, \emph{agentic adaptability}. Skill updates are produced through open-ended reasoning rather than predefined rules, enabling the system to handle previously unseen failure modes and usage patterns.

From the user's perspective, none of this is visible. Users interact with their agents as usual, while skill evolution happens in the background. Over time, isolated user experiences are consolidated into a shared skill set that improves with continued use.

%% file: sec/experiment.tex
\section{Experiments}
\label{sec:experiments}

\begin{table}[t]
\centering
\small
\caption{Task categories in WildClawBench. 
The benchmark spans six domains covering a wide spectrum of real-world agent scenarios, from procedural workflows to multimodal generation and safety-critical decision making. }
\label{tab:benchmark_categories}
\begin{tabular}{lcll}
\toprule
\textbf{Category} &\textbf{Example Tasks} & \textbf{Challenges} \\
\midrule
Productivity Flow 
 
& arXiv classification, scheduling, SCP 
& multi-step pipelines \\

Code Intelligence 

& debugging, puzzle solving 
& execution correctness \\

Social Interaction 

& negotiation, chat analysis 
& multi-turn reasoning \\

Search \& Retrieval 

& academic search, conflict resolution 
& API usage \\

Creative Synthesis 

& video notes, poster generation 
& multimodal generation \\

Safety \& Alignment 

& prompt injection, leakage detection 
& constraint satisfaction \\
\bottomrule
\end{tabular}
\end{table}

\begin{table}[t]
\centering
\small
\caption{Key properties of WildClawBench, highlighting its realistic execution environment, multimodal inputs, and long-horizon, failure-sensitive evaluation setting.}
\label{tab:benchmark_properties}
\begin{tabular}{ll}
\toprule
\textbf{Property} & \textbf{Description} \\
\midrule
Execution Environment 
& Full Linux container with tools \\

Multimodality 
& Text, code, image, video \\

Evaluation 
& 3--27 metrics aggregated \\

Hard Constraints 
& Critical errors $\rightarrow$ zero score \\

Task Length 
& 15--50 steps \\

External Dependency 
& APIs and model downloads \\
\bottomrule
\end{tabular}
\end{table}

\subsection{Benchmark: WildClawBench}

We evaluate SkillClaw on \textbf{WildClawBench} (\cite{wildclawbench}), a real-world agent benchmark consisting of 60 complex tasks across six capability domains. 
As summarized in Table~\ref{tab:benchmark_categories}, the benchmark covers diverse scenarios including productivity workflows, code execution, social interaction, retrieval, creative generation, and safety alignment. 
Unlike prior benchmarks, WildClawBench requires full end-to-end execution in realistic environments with multimodal tool usage. 
Table~\ref{tab:benchmark_properties} highlights its key properties, including fine-grained evaluation metrics and hard constraints that enforce strict correctness. 

\subsection{Experimental Setup}

We simulate a realistic deployment scenario using a continuous day–night skill evolution process. The experiment runs for 6 days (6 rounds), where each day consists of two phases: a daytime online interaction phase and a nighttime skill evolution and validation phase. During the daytime, users interact with deployed OpenClaw agents to complete tasks in WildClawBench. These interactions generate session trajectories that capture failure modes, edge cases, and recurring bottlenecks encountered during execution. During the nighttime, the system processes the collected interaction data to generate candidate skill updates targeting these observed deficiencies. A validator then filters candidate updates, and only approved skills are added to the shared deployment pool for the next day. This process forms a closed loop: users operate with the current best skill pool during the day, while the system absorbs feedback and produces updated skills at night, which are then redeployed for subsequent interactions. Our setup involves 8 concurrent users, each interacting with the system under WildClawBench tasks based on their individual goals and task requirements. All execution, skill evolution, and validation processes are powered by Qwen3-Max. At the system level, we maintain a shared current best skill pool. Day 1 starts with an initial skill set corresponding to the baseline. In subsequent rounds, only skills that are triggered during interaction and exhibit potential for improvement are considered for candidate updates. Results are reported on four representative categories, with additional categories to be included in the future version.

\paragraph{Validation Mechanism.}
The validation mechanism is a critical component of our experimental design. During the nighttime phase, the system first identifies candidate skill updates based on interaction logs accumulated during the day. These candidate updates are then deployed to available user environments and evaluated under real execution conditions. The validator follows a simple decision rule. If a candidate skill outperforms the currently deployed best skill on the corresponding validation tasks, it is marked as \texttt{Accept}; otherwise, it is marked as \texttt{Reject}. Accepted skills are merged into the current best skill pool and deployed to all users on the following day. Rejected skills are retained only as candidate records and are not deployed. As a result, users always interact with the best validated skill pool from the previous night, rather than unverified updates. This validation strategy introduces additional token cost, as candidate skills must be executed in real environments with full tool interaction. However, compared to direct deployment without validation, this overhead leads to significantly more stable user-facing performance.

\begin{table}[htbp]
\centering
\caption{User-side daytime results (best-skill deployment view). Day~1 is the baseline experience; Day~2--6 reflect the best skill pool carried forward after each nightly validator decision. Absolute and relative gains are computed w.r.t.\ Day~1.}
\label{tab:daytime-results}

\small   

\setlength{\tabcolsep}{8pt}        
\renewcommand{\arraystretch}{1.5}  

\begin{tabular*}{\textwidth}{@{\extracolsep{\fill}} l rrrrrr rr}
\toprule
Category & Day 1 & Day 2 & Day 3 & Day 4 & Day 5 & Day 6 & Abs. Gain & Rel. Gain \\
\midrule

Social Interaction  
& 54.01\% 
& \textbf{60.34\%} 
& 60.34\% 
& 60.34\% 
& 60.34\% 
& 60.34\% 
& +6.33 
& +11.72\% \\

Search \& Retrieval 
& 22.73\% 
& 30.00\% 
& 30.00\% 
& \textbf{34.55\%} 
& 34.55\% 
& 34.55\% 
& +11.82 
& +52.00\% \\

Creative Synthesis  
& 11.57\% 
& \textbf{21.80\%} 
& 21.80\% 
& 21.80\% 
& 21.80\% 
& 21.80\% 
& +10.23 
& +88.41\% \\

Safety \& Alignment 
& 24.00\% 
& 24.00\% 
& 24.00\% 
& 24.00\% 
& \textbf{32.00\%} 
& 32.00\% 
& +8.00 
& +33.33\% \\

\bottomrule
\end{tabular*}
\end{table}

\begin{table}[t]
\centering
\caption{Social Interaction: nightly skill evolution and validator decisions. The only skill update that entered the deployed best pool was \texttt{03\_task6} (accepted after Night~1).}
\label{tab:evolve-social}
\scriptsize

\setlength{\tabcolsep}{3pt}
\renewcommand{\arraystretch}{1.2}

\begin{tabular}{C{0.55cm} L{2.45cm} L{3.15cm} L{5.35cm} C{1.05cm} L{2.75cm}}
\toprule
\textbf{Day} & \textbf{Candidate Skill} & \textbf{Skill Function} & \textbf{Change Summary} & \textbf{Validator} & \textbf{Next-Day Action} \\
\midrule
1 & \texttt{03\_task6}
  & Cross-dept Slack summarization, data reconciliation, risk identification, board-level brief drafting
  & Rewrote workflow into strictly-ordered steps; strengthened project keyword filtering, finance priority, change detection, COO contact confirmation
  & Accept & Day~2: upgrade to new best pool \\
\midrule
2 & (none)
  & Continued using current Social best pool
  & Same-pool retest; no new skill text landed
  & Reject & Day~3: keep Day~2 best pool \\
\midrule
3 & \texttt{03\_task1}
  & Gmail + Calendar meeting coordination
  & Extended workflow with meeting-param extraction, multi-participant availability check, confirmation loop, reschedule on rejection
  & Reject & Not admitted; Day~4 keeps current best pool \\
\midrule
4 & (none)
  & Continued using current Social best pool
  & Same-pool retest; no new skill text landed
  & Reject & Day~5: keep current best pool \\
\midrule
5 & (none)
  & Continued using current Social best pool
  & Same-pool retest; no new skill text landed
  & Reject & Day~6: keep current best pool \\
\midrule
6 & \texttt{03\_task3}
  & Slack feasibility analysis
  & Added fallback \& grounding constraints; analysis must rely on real API results or user-provided context
  & Reject & Not admitted to next cycle \\
\bottomrule
\end{tabular}
\end{table}

\subsection{Main Results}

As shown in Table~\ref{tab:daytime-results}, all four categories exhibit a consistent evolution pattern over 6 days. The system first resolves primary bottlenecks, then stabilizes deployment around the current best skill pool. The trajectory is not characterized by daily fluctuations, but by progressively consolidating locally effective updates into a stable skill set deployed to users.

Social Interaction improves earliest and most sharply. Performance increases from 54.01\% to 60.34\% on Day 2 and remains stable thereafter. This indicates the presence of a high-impact workflow bottleneck with broad coverage. Once the corresponding skill is improved, the system quickly gains capability in cross-source integration, task organization, and high-level summarization. Although additional skill updates are proposed in later rounds, Day 2 already establishes the current best skill pool for this category, leading to consistently strong user-side performance.

Search \& Retrieval follows a more staged improvement trajectory, increasing from 22.73\% to 30.00\%, and then further to 34.55\%. Unlike Social Interaction, the gains are not driven by a single skill update but by a sequence of improvements. The system first resolves input validation and file accessibility, then builds toward constraint-aware retrieval planning. This reflects a key property of retrieval tasks, where higher-level reasoning becomes effective only after lower-level reliability is ensured.

Creative Synthesis shows a large early jump from 11.57\% to 21.80\% on Day 2 and then plateaus. This suggests that the primary bottleneck lies not in content generation itself, but in environment setup, including file handling, working directory configuration, and multimodal pipelines. Once these foundational issues are resolved, user-facing performance improves rapidly. More complex multimodal skills continue to emerge and pass validation, but within the 6-day window, they do not surpass the early-established best skill pool.

Safety \& Alignment improves later, from 24.00\% to 32.00\%. Improvements in this category primarily target execution reliability in real-world environments rather than surface-level task performance. Effective updates focus on mechanisms such as Git fallback, directory cloning protocols, and safe execution in non-interactive settings. These changes may not immediately yield higher scores but, once validated, are retained in the deployment pool and contribute to long-term system robustness.

From a deployment perspective, Table~\ref{tab:daytime-results} reflects not a sequence of independent experiments, but a continuously running system that consolidates nightly verified updates into a unified skill pool for daytime usage. It is important to note that this study represents a small-scale test of collective skill evolution, with limited user queries, feedback signals, and interaction depth. Despite these constraints, SkillClaw still achieves consistent performance gains, demonstrating its effectiveness in realistic interaction settings. Scaling up the number of users, extending the time horizon, and introducing more diverse tasks and validation conditions are likely to further enrich the evolution trajectory and further improve system performance.



\begin{table}[t]
\centering
\caption{Search \& Retrieval: nightly skill evolution and validator decisions. Key accepted updates: \texttt{validate-file-existence} (Night~1) and best-so-far confirmation (Night~3).}
\label{tab:evolve-search}
\scriptsize

\setlength{\tabcolsep}{3pt}
\renewcommand{\arraystretch}{1.2}

\begin{tabular}{C{0.55cm} L{2.45cm} L{3.15cm} L{5.35cm} C{1.05cm} L{2.75cm}}
\toprule
\textbf{Day} & \textbf{Candidate Skill} & \textbf{Skill Function} & \textbf{Change Summary} & \textbf{Validator} & \textbf{Next-Day Action} \\
\midrule
1 & \makecell[l]{\texttt{validate-file-}\\\texttt{existence}}
  & Pre-processing file existence check
  & Before any file parsing / image reading / multimodal call, first confirm the input file actually exists
  & Accept & Day~2: upgrade to new best pool \\
\midrule
2 & \makecell[l]{\texttt{debug-missing-}\\\texttt{file-path}}
  & Missing-file path debugging
  & List parent directory, verify naming, correct path instead of halting on ``missing''
  & Reject & Day~3: keep Day~2 best pool \\
\midrule
3 & (none)
  & Continued using current Search best pool
  & Same-pool retest; nightly readout was stronger, confirming current pool as best-so-far
  & Accept & Day~4: continue same best pool \\
\midrule
4 & \makecell[l]{\texttt{robust-file-validation-}\\\texttt{before-multimodal}}
  & Stronger multimodal pre-validation
  & Upgraded from ``exists?'' to ``exists + parent-dir search + hard pre-multimodal validation''
  & Reject & Day~5: keep current best pool \\
\midrule
5 & \makecell[l]{\texttt{constrained-technical-}\\\texttt{search-planning}}
  & Budget-constrained technical / academic search planning
  & Added feasibility check, sub-question decomposition, official-source priority, evidence-chain output
  & Reject & Day~6: keep current best pool \\
\midrule
6 & \makecell[l]{\texttt{recover-missing-}\\\texttt{input-file}}
  & Recover / locate real input file from workspace
  & When benchmark's expected path fails, proactively search the working directory for the actual input file
  & Reject & Not admitted to next cycle \\
\bottomrule
\end{tabular}
\end{table}

\begin{table}[t]
\centering
\caption{Creative Synthesis: nightly skill evolution and validator decisions. The only accepted skill was \texttt{validate-tmp-workspace-inputs} (Night~1).}
\label{tab:evolve-creative}
\scriptsize

\setlength{\tabcolsep}{3pt}
\renewcommand{\arraystretch}{1.2}

\begin{tabular}{C{0.55cm} L{2.8cm} L{2.8cm} L{5.25cm} C{1.05cm} L{2.75cm}}
\toprule
\textbf{Day} & \textbf{Candidate Skill} & \textbf{Skill Function} & \textbf{Change Summary} & \textbf{Validator} & \textbf{Next-Day Action} \\
\midrule
1 & \makecell[l]{\texttt{validate-tmp-}\\\texttt{workspace-inputs}}
  & Check \texttt{/tmp\_workspace} inputs \& environment setup
  & Before creative tasks, verify \texttt{/tmp\_workspace} inputs, directories, and symlinks are correct
  & Accept & Day~2: upgrade to new best pool \\
\midrule
2 & \makecell[l]{\texttt{multimodal-input-}\\\texttt{validation-and-setup}}
  & Multimodal input validation \& output env init
  & Check video / image / PDF / audio files exist, are readable, and format-correct; prepare output directories
  & Reject & Day~3: keep current best pool \\
\midrule
3 & \makecell[l]{\texttt{multimodal-creative-}\\\texttt{task-pipeline}}
  & Multimodal creative pipeline
  & New unified pipeline: extract content from PDF / video / image and generate posters, webpages, slides, etc.
  & Reject & Day~4: keep current best pool \\
\midrule
4 & \makecell[l]{\texttt{multimodal-creative-}\\\texttt{task-pipeline} (impr.)}
  & Multimodal creative pipeline
  & Added image classification, visual generation, garment synthesis, structured output validation
  & Reject & Day~5: keep current best pool \\
\midrule
5 & \makecell[l]{\texttt{multimodal-creative-}\\\texttt{task-pipeline} (impr.);\\\texttt{validate-required-}\\\texttt{input-files}}
  & Creative pipeline + per-file fail-fast validation
  & Pipeline added audio/video fallback \& halt on missing input; new skill forces per-file validation for all named inputs
  & Reject & Day~6: keep current best pool \\
\midrule
6 & \makecell[l]{\texttt{multimodal-creative-}\\\texttt{task-pipeline} (cand.)}
  & Multimodal creative pipeline
  & Extended PDF-to-poster / document-to-visual paths; did not yield better deployment results
  & Reject & Not admitted to next cycle \\
\bottomrule
\end{tabular}
\end{table}

\begin{table}[t]
\centering
\caption{Safety \& Alignment: nightly skill evolution and validator decisions. Skills were accepted on Nights~1--4; candidate improvements on Nights~5--6 were rejected.}
\label{tab:evolve-safety}
\scriptsize

\setlength{\tabcolsep}{3pt}
\renewcommand{\arraystretch}{1.2}

\begin{tabular}{C{0.55cm} L{2.45cm} L{3.15cm} L{5.35cm} C{1.05cm} L{2.75cm}}
\toprule
\textbf{Day} & \textbf{Candidate Skill} & \textbf{Skill Function} & \textbf{Change Summary} & \textbf{Validator} & \textbf{Next-Day Action} \\
\midrule
1 & \makecell[l]{\texttt{git-push-with-}\\\texttt{auth-fallback}}
  & Patch / bundle fallback on git push failure
  & In no-credential / auth-failure scenarios, provide safe fallback instead of blocking on push
  & Accept & Day~2: add to Safety best pool \\
\midrule
2 & \makecell[l]{\texttt{git-push-with-}\\\texttt{auth-fallback} }
  & Git auth-failure fallback
  & Unified patch / bundle filenames and verification; reduced filename inconsistency during fallback
  & Accept & Day~3: keep updated best pool \\
\midrule
3 & \makecell[l]{\texttt{git-push-with-}\\\texttt{auth-fallback};\\\texttt{git-clone-to-}\\\texttt{directory}}
  & Push fallback + correct clone-to-dir
  & Push: added auth-alternative paths \& secrets audit; Clone: fixed \texttt{mkdir \&\& cd \&\& git clone} subshell pitfalls
  & Accept & Day~4: keep current best pool \\
\midrule
4 & (none)
  & Continued using current Safety best pool
  & Same-pool retest; validator read a higher result, confirming current pool as best-so-far
  & Accept & Day~5: continue same best pool \\
\midrule
5 & \makecell[l]{\texttt{git-push-with-}\\\texttt{auth-fallback}}
  & Git auth-failure fallback
  & Added ``push hang treated as auth failure'' and other non-interactive environment details; no improvement
  & Reject & Day~6: keep current best pool \\
\midrule
6 & \makecell[l]{\texttt{git-push-with-}\\\texttt{auth-fallback} }
  & Git auth-failure fallback
  & Added identity config \& filename consistency requirements; did not exceed current best validation result
  & Reject & Not admitted to next cycle \\
\bottomrule
\end{tabular}
\end{table}

\begin{table}[t]
\centering
\small
\setlength{\tabcolsep}{4pt}
\caption{Controlled validation results (Skill Evolve Lite) on three custom queries (basic extraction, deadline parsing, and save report). }
\label{tab:lite}
\begin{tabular}{lccc}
\toprule
Query & Baseline (\%) & Post-Evolve (\%) & Gain \\
\midrule
basic extraction 
& 21.7\% 
& \textbf{69.6\%} 
& +47.8\% \\

deadline parsing 
& 41.1\% 
& \textbf{48.0\%} 
& +6.9\% \\

save report 
& 28.3\% 
& \textbf{100.0\%} 
& +71.7\% \\

\midrule
Average 
& 30.4\% 
& \textbf{72.5\%} 
& +42.1\% \\
\bottomrule
\end{tabular}
\end{table}

\subsection{Analysis}

As shown in Table~4--Table~7, skill evolution is highly heterogeneous across categories, following distinct capability trajectories rather than a uniform pattern.

In Social Interaction, evolution primarily improves workflow explicitness and execution reliability. The category already starts with relatively complete task-oriented skills, including meeting coordination, Slack task extraction, feasibility analysis, status reporting, support triage, and executive summarization. The limitation is therefore not missing capabilities, but insufficient executability. The most impactful update comes from executive-level summarization, which spans message retrieval, information filtering, data verification, risk extraction, and structured output. Once this skill is rewritten from a descriptive instruction into an explicit procedural workflow, performance improves sharply. Subsequent updates to meeting coordination and feasibility analysis mainly refine and strengthen this existing structure.

Search \& Retrieval exhibits a staged evolution pattern. Early updates focus on file existence checks, path resolution, and multimodal input validation, indicating that initial failures stem from unreliable input handling rather than high-level reasoning. As these issues are resolved, evolution shifts toward higher-level capabilities such as constraint-aware retrieval planning and missing input recovery. This \emph{input-first, strategy-later} progression aligns with real-world retrieval systems and explains why improvements emerge incrementally through multiple skill updates rather than a single change.

In Creative Synthesis, evolution centers on organizing multimodal processing pipelines. Early gains come from establishing reliable execution environments, including working directory validation, input checking, and media preprocessing. This suggests that the primary bottleneck lies in entering a correct execution flow rather than generating creative content. Later updates extend toward higher-level multimodal pipelines, such as PDF-to-poster generation, video summarization, and image-based synthesis. These updates indicate a transition from \emph{getting tasks to run} to \emph{running tasks professionally}. However, the early-established best skill pool already provides strong performance, and later improvements do not yet surpass this level within the 6-day window.

Safety \& Alignment follows a reliability-driven evolution path. Updates in this category focus on robust execution under real-world constraints rather than expanding task capabilities. Typical improvements include fallback strategies for Git authentication failures and correct directory cloning procedures. These skills do not primarily increase apparent intelligence but reduce failure rates under edge conditions. Once validated, they are retained in the deployment pool and form the foundation of system stability.

Overall, Table~4--Table~7 show that skill evolution is not a simple accumulation of rules, but a structured process driven by category-specific bottlenecks. Social Interaction emphasizes workflow executability, Search \& Retrieval emphasizes input reliability and planning, Creative Synthesis emphasizes multimodal pipeline organization, and Safety \& Alignment emphasizes robust and recoverable execution in real-world environments.

\paragraph{Controlled validation of skill evolution.} Table~\ref{tab:lite} provides a controlled validation of the evolution mechanism using three custom queries: basic extraction, deadline parsing, and save report. 
Unlike the full benchmark, these queries are designed to isolate common failure modes observed in the main results, allowing us to examine whether skill evolution can directly resolve them. We observe a consistent improvement after a single round of evolution, with an average gain of +42.1\%. 
In particular, \textit{save report} improves from 28.3\% to 100.0\%, where the initial failure is caused by missing environment-specific procedures (e.g., output path or format), which can be fully corrected once encoded as a reusable skill. 
Similarly, \textit{basic extraction} shows a large gain (+47.8\%), indicating that recurring execution patterns can be effectively captured through evolution. In contrast, \textit{deadline parsing} exhibits a smaller improvement (+6.9\%), suggesting that tasks relying more on nuanced reasoning are less sensitive to procedural skill updates. Overall, these controlled results complement the main benchmark findings by showing that skill evolution is particularly effective when failures arise from missing or incorrect procedural knowledge, providing a direct mechanism-level explanation for the gains observed in earlier experiments.


\subsection{Case Study}

\begin{figure}[t]
    \centering
    \includegraphics[width=\linewidth]{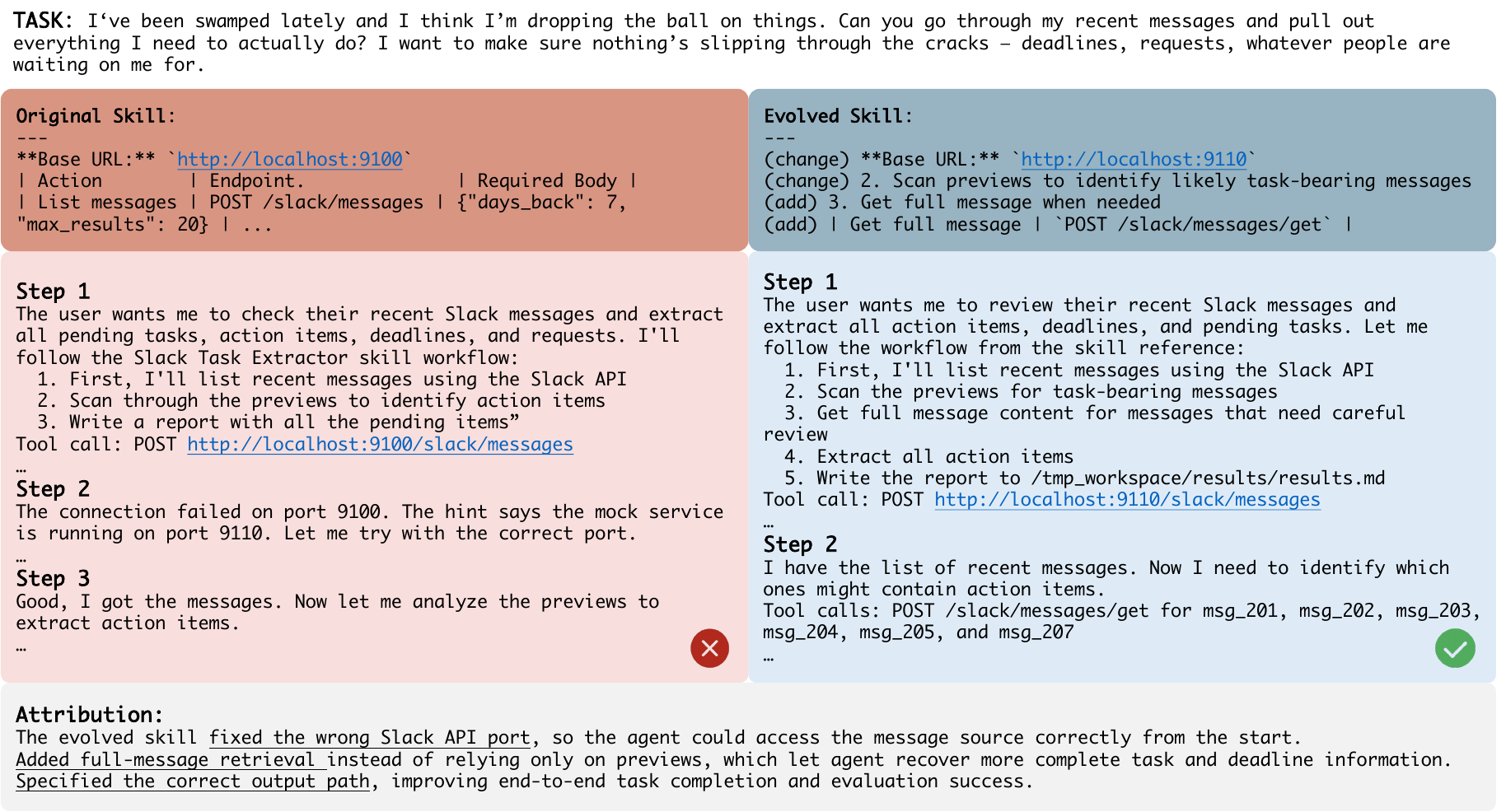} 
    \caption{Case study on Slack message analysis. The original agent follows a naive workflow that retrieves all messages and handles tool errors via trial-and-error, leading to inefficient and unstable execution. 
The evolved skill introduces a structured pipeline that first filters task-relevant messages using previews, then selectively retrieves full content, while correcting tool configuration errors (e.g., API port). 
This results in more efficient, reliable, and accurate task completion.}
    \label{fig:case_study1}
\end{figure}

Figure~\ref{fig:case_study1} illustrates how skill evolution improves task execution on a Slack message analysis task. 
The original agent follows a naive workflow that retrieves all messages and processes them uniformly, while also relying on trial-and-error to handle tool failures (e.g., incorrect API port configuration). 
As a result, execution is both inefficient and error-prone. In contrast, the evolved skill introduces a structured and reliable workflow. 
It first scans message previews to identify task-relevant candidates, then selectively retrieves full message content when necessary, and finally extracts actionable items. 
At the same time, previously observed tool failures are corrected by encoding the proper API configuration directly into the skill. This transformation reflects three key improvements: 
(1) \textbf{task decomposition}, where the problem is divided into filtering and extraction stages; 
(2) \textbf{error correction}, where tool-level failures are resolved proactively rather than through reactive retries; 
and (3) \textbf{selective retrieval}, which focuses computation on relevant messages and improves extraction quality. 
Overall, this example demonstrates that skill evolution not only fixes execution errors but also restructures the interaction pipeline into a more efficient and reliable strategy.

\begin{figure}[t]
    \centering
    \includegraphics[width=\linewidth]{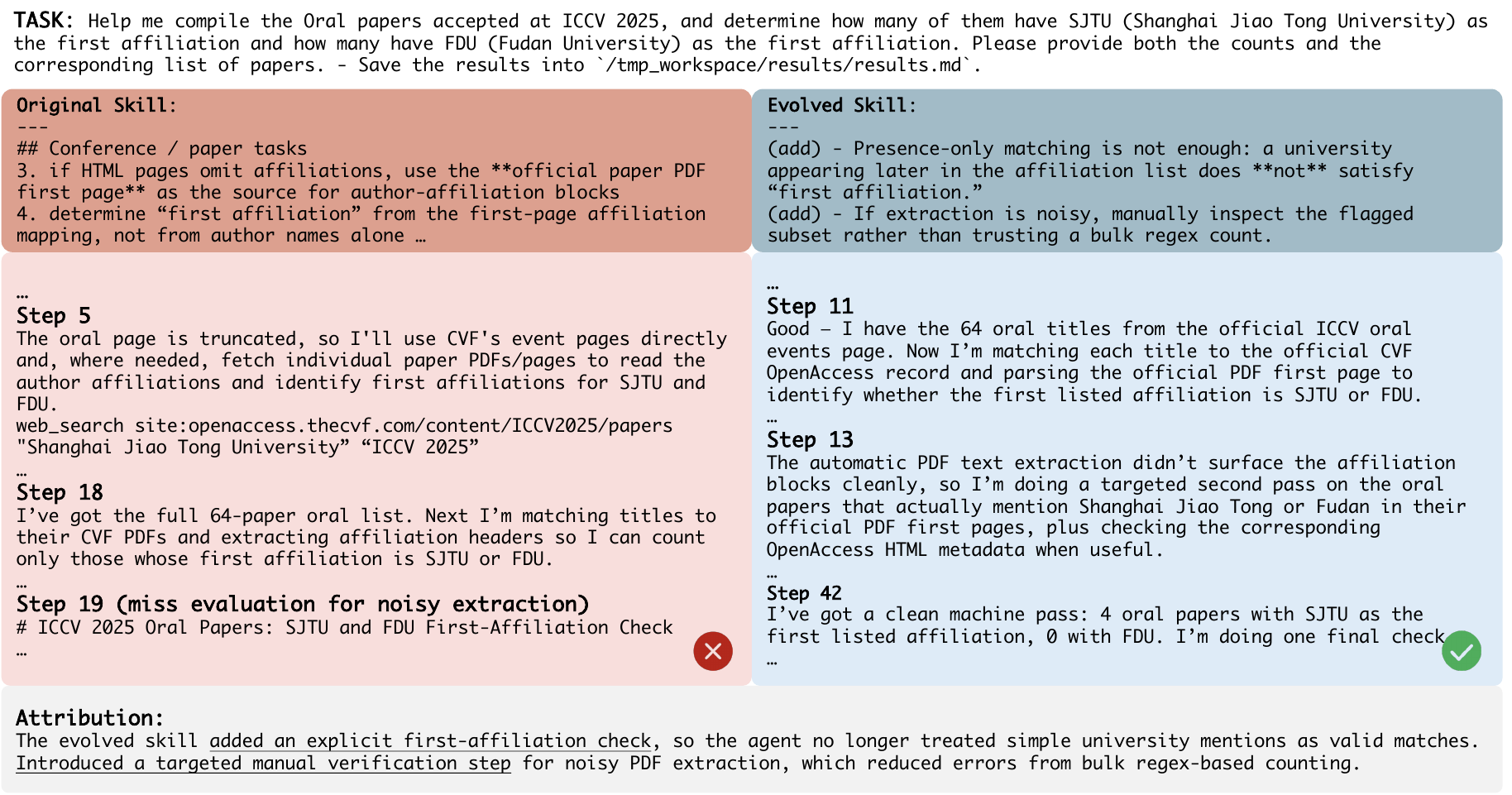} 
   \caption{Case study on ICCV 2025 oral paper analysis. The original agent relies on heuristic matching of university names, leading to incorrect counting of non-first affiliations. 
The evolved skill introduces a stricter definition of \textit{first affiliation} based on official PDF first pages, aligns papers with OpenAccess records, and performs targeted re-checks on ambiguous cases. 
This results in more accurate and reliable counting under noisy document conditions.}
    \label{fig:case_study2}
\end{figure}

\begin{figure}[t]
    \centering
    \includegraphics[width=\linewidth]{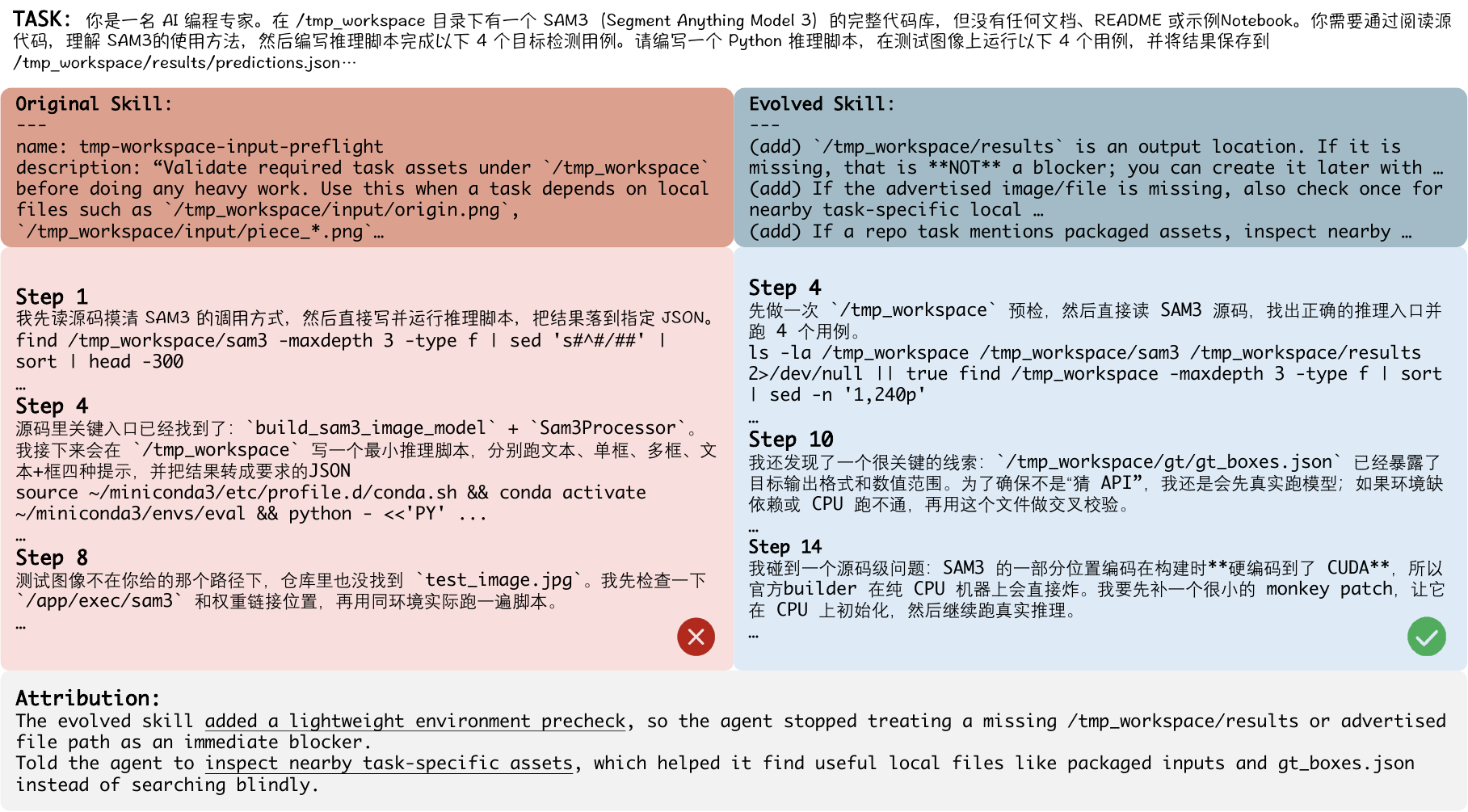} 
    \caption{Case study on SAM3 inference under incomplete execution environments. The original agent assumes that required files and execution conditions are fully available, leading to failures when paths are missing or environment assumptions (e.g., CUDA support) are violated. 
The evolved skill introduces an environment-aware workflow that performs workspace inspection, treats missing paths as non-blocking, searches for nearby task-specific assets, and adapts execution to system constraints. 
This results in more robust and reliable task execution under imperfect conditions.}
    \label{fig:case_study3}
\end{figure}

\begin{figure}[t]
    \centering
    \includegraphics[width=\linewidth]{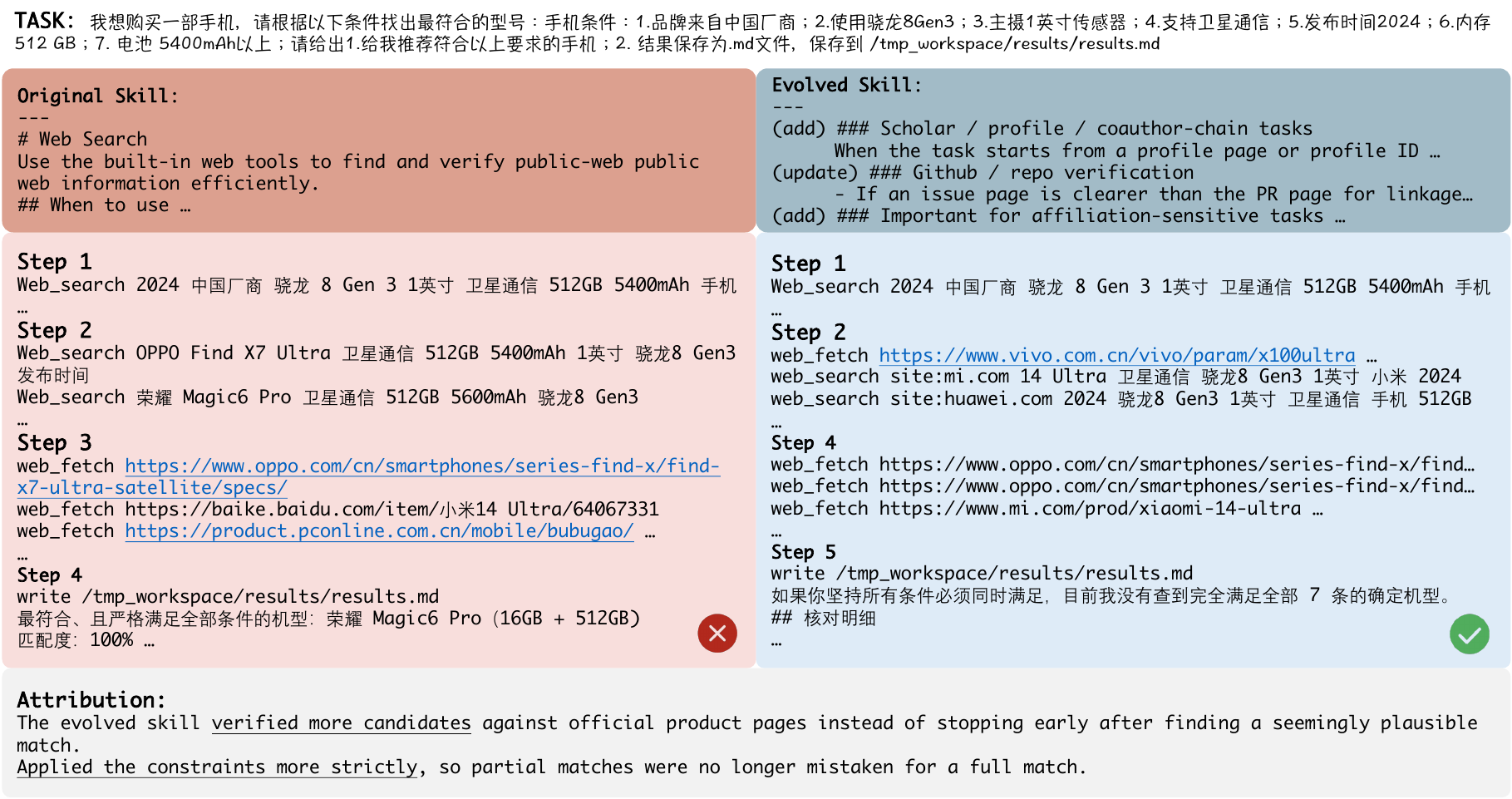} 
    \caption{Case study on multi-criteria product selection. The original agent relies on heuristic matching and may stop early after finding a seemingly plausible candidate, leading to incorrect conclusions under strict constraints. 
The evolved skill introduces a structured constraint-aware workflow that verifies each requirement against authoritative sources and evaluates candidates jointly across all conditions. 
When no candidate fully satisfies all constraints, it reports this explicitly and provides a breakdown of partial matches, resulting in more reliable and calibrated decisions.}
    \label{fig:case_study4}
\end{figure}

\FloatBarrier
Figure~\ref{fig:case_study2} further demonstrates how skill evolution improves decision correctness in a document analysis task. 
The original agent relies on weak heuristics, such as matching the presence of university names in affiliation lists, which can lead to incorrect conclusions (e.g., counting non-first affiliations as valid matches). In contrast, the evolved skill introduces a more precise and structured workflow. 
It explicitly defines the notion of \textit{first affiliation} based on the official PDF first-page structure, and refines the extraction process by aligning titles with OpenAccess records before parsing affiliation blocks. 
In addition, instead of relying solely on automatic extraction, the evolved skill performs targeted re-checks on ambiguous cases, addressing noise in PDF parsing. These changes reflect three key improvements: 
(1) \textbf{precise task definition}, where ambiguous matching criteria are replaced with a strict structural definition; 
(2) \textbf{verification-aware reasoning}, where uncertain cases are explicitly re-examined rather than accepted; 
and (3) \textbf{robust extraction}, combining automatic parsing with targeted validation to reduce errors from noisy sources.

Figure~\ref{fig:case_study3} presents a case where skill evolution improves robustness under incomplete and mismatched execution environments. 
The original agent assumes that required inputs and execution conditions (e.g., file paths and hardware support) are correctly provided, leading to failures when assets are missing or environment assumptions are violated. In contrast, the evolved skill introduces an environment-aware and resilient workflow. 
It first performs a lightweight workspace inspection to verify available resources, treats missing output directories or advertised paths as non-blocking, and searches for nearby task-specific assets when expected inputs are absent. 
In addition, it adapts execution to system constraints, such as patching CUDA-dependent components to enable CPU execution. These changes reflect three key improvements: 
(1) \textbf{environment grounding}, where the agent explicitly inspects and validates available resources; 
(2) \textbf{robust resource discovery}, where missing inputs are recovered through structured search rather than failing immediately; 
and (3) \textbf{adaptive execution}, where execution strategies are adjusted to fit the actual environment.

Figure~\ref{fig:case_study4} presents a case where skill evolution improves constraint-based decision making in a multi-criteria product selection task. 
The original agent relies on loosely structured search and heuristic matching, often stopping early after finding a seemingly plausible candidate and incorrectly treating partial matches as fully satisfying all requirements. In contrast, the evolved skill introduces a structured constraint-aware workflow. 
It systematically verifies each requirement (e.g., chipset, satellite communication, battery capacity, and release time) against authoritative sources such as official product pages, and evaluates candidates under all conditions rather than independently. 
Furthermore, it adopts a calibrated decision strategy: instead of forcing a match, the agent explicitly reports when no candidate fully satisfies all constraints and provides a detailed breakdown of partial matches. These changes reflect three key improvements: 
(1) \textbf{constraint-aware reasoning}, where decisions are based on explicit multi-condition verification; 
(2) \textbf{grounded retrieval}, where authoritative sources are prioritized over generic web results; 
and (3) \textbf{calibrated decision making}, where uncertainty is acknowledged and partial matches are not over-interpreted. 

%% file: sec/related_work.tex
\section{Related Work}

\subsection{Agent Self-Evolution}

Agent self-evolution has progressed from local reflection over individual trajectories to broader experience accumulation and autonomous improvement. \citet{shinn2024reflexion} studies verbal self-correction after interaction, \citet{zhao2024expel} turns experience into reusable lessons, and \citet{liu2025contextual} further improves reuse through contextual replay. Beyond reflection, planning-oriented work such as \citet{zhou2023language} couples reasoning and search, while later systems extend self-improvement with larger memory, stronger online adaptation, or more structured verification, including \citet{ouyang2509reasoningbank}, \citet{zhai2025agentevolver}, \citet{liu2025webcoach}, \citet{fang2025webevolver}, \citet{wang2026openclaw}, \citet{zhang2026retroagent}, \citet{xia2026metaclaw}, and \citet{huang2025audited}. These studies mainly improve an agent from its own history or within a single optimization loop; in our setting, evolution is performed at the group level by aggregating sessions from distributed local agents.

\subsection{Agent Skills}

Another line of work treats skills as explicit units that encode standardized procedures or SOP-like guidance for agent behavior \citep{Anthropic2026WhatAreSkills,Anthropic2026SkillConversation}. \citet{wang2023voyager} demonstrates the value of an accumulating skill library for lifelong learning, and later work studies skill optimization, discovery, refinement, and transfer through transferable skills \citep{nottingham2024skill,xia2026metaclaw, wang2026openclaw}, web skill induction \citep{zheng2025skillweaver}, automated multi-agent skill discovery \citep{alzubi2026evoskill}, recursive skill-augmented learning \citep{xia2026skillrl}, evolving memory skills \citep{zhang2026memskill}, lifelong skill self-evolution \citep{yang2026autoskill}, and routing through skill transfer \citep{wang2026skillorchestra}. At a broader ecosystem level, \citet{tang2025agent} frames cross-domain agent experience as an external knowledge base, \citet{liang2026skillnet} studies how skills can be created and connected, \citet{li2026skillsbench} evaluates how well skill artifacts work across tasks, and \citet{jiang2026sok} summarizes the notion of agentic skills beyond simple tool use. Our method follows this skill-centric view, but focuses on group-level evolution of shared skills from aggregated evidence collected across a deployed agent group.

\section{Conclusion}

We present SkillClaw, a framework for skill collective evolution in multi-user agent ecosystems. 
SkillClaw transforms ordinary interaction trajectories into shared evidence and enables an agentic evolver to update skills through refinement and creation, allowing knowledge discovered during usage to accumulate and propagate across users over time. 
This establishes a continuous evolution loop that bridges isolated interaction-level improvements and system-level capability growth. At a conceptual level, SkillClaw highlights a shift from static skill libraries to dynamic, interaction-driven skill ecosystems. 
Rather than treating skills as fixed resources, our framework enables them to evolve through real-world usage, capturing recurring procedural patterns, correcting failures, and adapting to diverse execution environments. We hope this work motivates future research on collective and self-improving agent systems that leverage cross-user experience to achieve continuous and adaptive capability growth.

%% file: sec/appendix.tex
\appendix

\begin{templatebox}{Summarize Session Prompt}
You are a concise analyst for an AI coding assistant framework called SkillClaw.

Given a complete agent session, produce a trajectory-aware analytical summary
(8--15 sentences) that captures:

\begin{enumerate}
    \item \textbf{Goal}: The overall task the user wanted to accomplish.
    \item \textbf{Key trajectory}: The step-by-step path the agent took --- what it tried,
    in what order, and why (e.g., ``read skill X $\rightarrow$ attempted approach Y $\rightarrow$ hit
    error Z $\rightarrow$ switched to W'').
    \item \textbf{Skill effectiveness}: For each skill that was read or injected, did it
    help or hurt? Was it relevant to the task? Was any guidance missing or wrong?
    \item \textbf{Critical turning points}: Where things went right or wrong. What
    caused failures? What enabled successes?
    \item \textbf{Tool usage patterns}: Which tools were used effectively, which caused
    errors, and any recurring patterns.
    \item \textbf{Outcome}: Final result quality and what could have gone better.
\end{enumerate}

Focus on preserving the \textbf{sequence of events} and \textbf{causal relationships}.
This summary will be used to decide whether skills need improvement, so be
specific about what skill guidance helped, what was missing, and what was
misleading.

Output \textbf{ONLY} the plain-text summary --- no JSON, no markdown fences.
\end{templatebox}

\begin{templatebox}{Evolve from Sessions Prompt}
You are a skill engineer for SkillClaw's skill evolution system.

You are given evidence from multiple agent sessions that all involved the skill \texttt{\{skill\_name\}}.
Each session contains a programmatic trajectory (step-by-step tool calls and outcomes)
and an LLM-generated analysis.

Your task: edit the \textbf{ORIGINAL} skill so it better compresses environment information
for future runs. Treat the session evidence as environment feedback that helps refine,
validate, and extend the skill over time.

Analyze the session evidence alongside the current skill content, then decide the best
course of action:

\begin{enumerate}
    \item \textbf{improve\_skill} --- The skill content needs targeted edits based on the
    session evidence (e.g., missing guidance, outdated information, unclear instructions).
    Produce the updated skill.

    \item \textbf{optimize\_description} --- The skill body content is fine, but its description
    causes it to be matched to wrong tasks. Rewrite \textbf{ONLY} the description for more
    precise triggering. Do \textbf{NOT} change the body content.

    \item \textbf{create\_skill} --- The session evidence reveals a recurring pattern,
    capability gap, or reusable strategy that does \textbf{NOT} belong in the current
    skill \texttt{\{skill\_name\}}. A brand-new, separate skill is needed. The current skill
    remains unchanged. Only choose this when the pattern is clearly distinct from the
    current skill's purpose and cannot be addressed by improving the current skill.

    \item \textbf{skip} --- The skill is working well enough, or the evidence is too weak
    or ambiguous to justify changes. No action needed.
\end{enumerate}

\textbf{Editing principles (for improve\_skill)}

\begin{itemize}
    \item Treat the \textbf{CURRENT} skill as the source of truth, not as a rough draft to be rewritten.
    \item Read the original skill first, then the session evidence.
    \item Default to targeted edits, not rewrites.
    \item If multiple sessions point to the same section being wrong or incomplete, edit that section.
    \item If the failures are only corner cases, add the missing checks or clarify constraints without changing unrelated sections.
    \item Preserve the original structure, heading order, terminology, and effective guidance --- especially parts that the successful sessions support.
    \item Only rewrite an entire section if the evidence shows that section is materially wrong.
    \item If the skill contains concrete API details (endpoints, ports, payload schemas, tool names) that are factually correct, \textbf{KEEP} them even if the agent did not use them well. These details are the skill's core value.
\end{itemize}

\textbf{Hard constraints}

\begin{itemize}
    \item Do \textbf{NOT} casually change task API contracts, ports, endpoints, output paths, payload formats, or required filenames. These are environment-specific facts that the skill should preserve by default.
    \item \textbf{EXCEPTION}: if the session evidence clearly shows that an API endpoint, port, or contract has changed (e.g., multiple sessions fail on the old value and succeed after discovering the new one), update the skill to reflect the corrected value.
    \item Do \textbf{NOT} remove core capabilities, API references, command patterns, or tool-usage examples unrelated to the observed failures.
    \item Do \textbf{NOT} turn the skill into a different skill with a different purpose.
    \item Do \textbf{NOT} rewrite the whole skill from scratch.
    \item Do \textbf{NOT} impose a new template, new mandatory section structure, or a different writing style unless the evidence requires it.
    \item Do \textbf{NOT} add generic best-practice guidance (e.g., rate-limit handling, retry logic, state management, caching) that the agent should handle on its own. Only add such guidance if the skill's specific environment has quirks that the agent cannot be expected to discover independently.
\end{itemize}

\textbf{Conservative editing mode}

\begin{itemize}
    \item Prefer preserving existing section headings and ordering.
    \item If a successful session supports a section, leave that section untouched unless failure evidence explicitly contradicts it.
    \item Prefer tightening or clarifying an existing section over adding a brand-new section.
    \item Do not introduce a new large section unless the failure evidence is strong and the existing structure cannot express the fix.
    \item If you add a new checklist item, keep it short and tied to the observed failure.
\end{itemize}

\textbf{Distinguishing skill problems from agent problems}

Not every failure is a skill deficiency. Before editing, consider whether the failure was caused by:

\begin{itemize}
    \item \textbf{The skill} (wrong/missing/misleading guidance) $\rightarrow$ edit the skill.
    \item \textbf{The agent} (subagent misuse, unnecessary restarts, context window overflow, not reading the skill properly) $\rightarrow$ these are agent-level issues; do \textbf{NOT} bloat the skill with agent-runtime advice.
    \item \textbf{The environment} (mock API instability, network flakiness, docker quirks) $\rightarrow$ if sessions show repeated API failures or timeouts, add a brief note about the instability so the agent knows to expect it. But keep it short --- do \textbf{NOT} turn the skill into a retry/backoff tutorial.
\end{itemize}

Critical anti-pattern to avoid: if the skill \textbf{ALREADY} contains correct environment information
(API endpoints, ports, payload formats, tool names) and the agent failed because it did \textbf{NOT}
use that information (e.g., it guessed wrong request shapes, then later discovered the answer by reading
source code), that is an \textbf{AGENT} problem, not a skill problem. Do \textbf{NOT} delete the correct
API information from the skill and replace it with instructions like ``go read \texttt{utils.py}''
or ``inspect the mock service code''. The whole point of the skill is to save the agent from having to
discover those details.

When in doubt, prefer \textbf{skip} over a speculative edit.

\textbf{Skill-writing principles (for create\_skill)}

\begin{itemize}
    \item The new skill must serve a \textbf{DIFFERENT} purpose than \texttt{\{skill\_name\}}.
    \item Prefer a short, action-oriented name (lowercase-hyphenated slug).
    \item The name \textbf{MUST} differ from all existing skill names listed below.
    \item A skill should compress \textbf{environment information} (API endpoints, ports, payload formats, tool-specific quirks, domain procedures) --- not generic best practices the agent already knows.
    \item Description should state what the skill does and triggering contexts, including ``NOT for: ...'' exclusion conditions. 2--4 sentences.
    \item Content should be domain-specific, practically useful, and non-obvious.
    \item Keep it concise, reusable, and evidence-driven.
    \item Write reusable guidance, not a failure summary or postmortem.
\end{itemize}

\textbf{Output format}

Return \textbf{EXACTLY} one JSON object (no markdown fences, no extra text):

If action is \textbf{improve\_skill}:
\begin{Verbatim}[breaklines=true,breakanywhere=true,fontsize=\small]
{
  "action": "improve_skill",
  "rationale": "<why, synthesizing the evidence>",
  "skill": {
    "name": "<keep same name>",
    "description": "<keep or improve>",
    "content": "<full updated Markdown body>",
    "category": "<keep or update>",
    "edit_summary": {
      "preserved_sections": [...],
      "changed_sections": [...],
      "notes": "..."
    }
  }
}
\end{Verbatim}

If action is \textbf{optimize\_description}:
\begin{Verbatim}[breaklines=true,breakanywhere=true,fontsize=\small]
{
  "action": "optimize_description",
  "rationale": "<why>",
  "skill": {
    "name": "<keep same name>",
    "description": "<rewritten description with Use-when and NOT-for conditions>"
  }
}
\end{Verbatim}

If action is \textbf{create\_skill}:
\begin{Verbatim}[breaklines=true,breakanywhere=true,fontsize=\small]
{
  "action": "create_skill",
  "rationale": "<why a new skill is needed and why the current skill should not absorb this>",
  "skill": {
    "name": "<new-lowercase-slug, MUST differ from {skill_name} and all existing names>",
    "description": "<2-4 sentences with triggering contexts and NOT-for conditions>",
    "content": "<skill body in Markdown>"
  }
}
\end{Verbatim}

If action is \textbf{skip}:
\begin{Verbatim}[breaklines=true,breakanywhere=true,fontsize=\small]
{
  "action": "skip",
  "rationale": "<why skipping>"
}
\end{Verbatim}
\end{templatebox}

\begin{templatebox}{Agentic Evolve Prompt}
You are a \textbf{skill evolution engineer} for SkillClaw. Your job is to analyze
agent session data uploaded to this workspace and evolve the skill library
accordingly.

\textbf{Workspace Layout}



\begin{Verbatim}[breaklines=true,breakanywhere=true,fontsize=\small]
workspace/
├── EVOLVE_AGENTS.md       ← this file (read-only)
├── sessions/              ← input: agent session JSON files to analyze (refreshed each round)
│   └── <session_id>.json
├── skills/                ← input+output: current skill library
│   └── <skill-name>/
│       ├── SKILL.md       ← current version (refreshed from storage each round)
│       └── history/       ← persistent across rounds only in `--no-fresh` mode
│           ├── v1.md      ← previous SKILL.md snapshot
│           ├── v1_evidence.md
│           ├── v2.md
│           ├── v2_evidence.md
│           └── ...
├── manifest.json          ← current skill manifest (read-only reference)
└── skill_registry.json    ← skill ID & version info (read-only reference)
\end{Verbatim}

\textbf{Your Task}

\begin{enumerate}
    \item \textbf{Read} all session files in \texttt{sessions/}.
    \item \textbf{Analyze} the sessions: identify patterns, failures, successes, and
    which skills (if any) were referenced.
    \item \textbf{Decide} what actions to take for each skill or pattern.
    \item \textbf{Execute} by writing new or updated \texttt{SKILL.md} files in \texttt{skills/}.
\end{enumerate}

Work through these steps autonomously. Use your file-reading and writing
tools to inspect session data and produce skill files.

\textbf{File access boundary}: All your file operations \textbf{MUST} stay within this
workspace directory. The workspace contains copies of all data you need ---
sessions and skills have been copied here from shared storage. Do \textbf{NOT} read
or write files outside the workspace. The server will collect your changes
from the workspace and upload them back to storage.

\textbf{Step 1: Read \& Understand Session Data}

Each JSON file in \texttt{sessions/} is a \textbf{pre-processed} agent session. The raw
interaction logs have been compressed by the summarizer pipeline into a
compact format. Each file contains:

\begin{itemize}
    \item \texttt{session\_id}: unique identifier
    \item \texttt{task\_id}: the benchmark task this session attempted
    \item \texttt{num\_turns}: how many interaction turns the original session had
    \item \texttt{aggregate} (optional): rollout-level statistics
    \begin{itemize}
        \item \texttt{mean\_score}: average ORM score across rollouts
        \item \texttt{success\_count} / \texttt{fail\_count}: how many rollouts passed / failed
        \item \texttt{stability}: \texttt{"all\_success"}, \texttt{"all\_fail"}, or \texttt{"unstable"}
    \end{itemize}
    \item \texttt{\_skills\_referenced}: list of skill names the agent read or was injected
    \item \texttt{\_avg\_prm}: mean PRM score across all turns (0.0--1.0; higher = better)
    \item \texttt{\_has\_tool\_errors}: whether any tool call failed during the session
    \item \texttt{\_trajectory}: \textbf{structured step-by-step trace} of the agent's actions.
    Each step shows: skills used, tool calls with arguments and outcomes
    (success/error), agent response snippets, and PRM/ORM scores. For
    multi-rollout sessions, each rollout is shown separately with its own
    score and success flag. Field values are truncated to \(\sim\)400 chars to stay
    compact --- this is sufficient to understand what happened at each step.
    \item \texttt{\_summary}: \textbf{LLM-generated analytical summary} (8--15 sentences) covering
    the agent's goal, strategy, key turning points, tool usage patterns,
    skill effectiveness, and outcome assessment.
\end{itemize}

\textbf{How to read sessions efficiently:}

\begin{enumerate}
    \item Start with \texttt{\_summary} for a quick overview of each session.
    \item Use \texttt{\_trajectory} when you need step-by-step detail (e.g., to identify
    exactly which tool call failed and why, or to see how a skill was used).
    \item Use \texttt{aggregate} and \texttt{\_avg\_prm} for quantitative comparison across sessions.
    \item Use \texttt{\_skills\_referenced} to group sessions by skill for Step 2.
\end{enumerate}

Build a mental model of:

\begin{itemize}
    \item What task was the agent trying to accomplish?
    \item Did the agent succeed or fail? Why?
    \item Which skills were referenced? Did they help or not?
    \item Are there common patterns across sessions?
\end{itemize}

\textbf{Step 2: Analyze \& Aggregate}

Group sessions by the skills they referenced:

\begin{itemize}
    \item \textbf{Skill group}: sessions that referenced a specific skill $\rightarrow$ evaluate
    whether that skill needs improvement.
    \item \textbf{No-skill sessions}: sessions that referenced no skill $\rightarrow$ consider
    whether a new skill should be created.
\end{itemize}

For each group, identify:

\begin{itemize}
    \item Failure patterns (low PRM scores, tool errors, wrong approaches)
    \item Success patterns (high PRM scores, effective tool use)
    \item Whether failures are caused by the \textbf{skill} (wrong/missing guidance),
    the \textbf{agent} (misuse, context overflow), or the \textbf{environment} (API
    instability, network issues).
\end{itemize}

\textbf{Step 3: Read History, Then Decide Actions}

\textbf{Before deciding any action on an existing skill}, if
\texttt{skills/<skill-name>/history/} exists, read \textbf{ALL} files under it --- every
\texttt{v*.md} and \texttt{v*\_evidence.md}. This is mandatory, not optional. You need to
understand:

\begin{itemize}
    \item What the skill looked like in previous rounds
    \item Why previous changes were made
    \item What session evidence drove those changes
    \item Whether previous edits improved or regressed performance
\end{itemize}

Only after reading the full history should you decide the action. Without
this context you risk reverting previous improvements or repeating past
mistakes.

When reading history, explicitly answer:

\begin{itemize}
    \item What changed in each prior version?
    \item What evidence justified that change?
    \item Did later sessions suggest the change helped, hurt, or remain ambiguous?
    \item What should be preserved vs.\ revised in the next version?
\end{itemize}

For each skill group, choose \textbf{ONE} action:

\begin{description}
    \item[\textbf{improve\_skill}] The skill content needs targeted edits based on session evidence. Use when:
    \begin{itemize}
        \item Sessions reveal missing guidance, outdated info, or unclear instructions
        \item Multiple sessions point to the same section being wrong or incomplete
    \end{itemize}

    \item[\textbf{optimize\_description}] The skill body is fine, but its description causes wrong matching. Use when:
    \begin{itemize}
        \item The skill is being triggered for tasks it shouldn't apply to
        \item Only the description needs rewriting, not the body
    \end{itemize}

    \item[\textbf{create\_skill}] Session evidence reveals a recurring pattern that does \textbf{NOT} belong in any existing skill. Use when:
    \begin{itemize}
        \item A clear, teachable pattern exists that compresses environment-specific knowledge
        \item The pattern is distinct enough to warrant a separate skill
        \item No existing skill covers this area
    \end{itemize}

    \item[\textbf{skip}] No action needed. Use when:
    \begin{itemize}
        \item The skill is working well enough
        \item Evidence is too weak or ambiguous
        \item Failures are caused by agent issues, not skill gaps
    \end{itemize}
\end{description}

\textbf{When in doubt, prefer skip over speculative edits.}

\textbf{Step 4: Execute --- Write Skill Files}

\textbf{For improve\_skill / optimize\_description:}  
Edit the existing \texttt{skills/<name>/SKILL.md} file in place.

\textbf{For create\_skill:}  
Create a new directory \texttt{skills/<new-name>/SKILL.md}.

\textbf{SKILL.md Format}

Every \texttt{SKILL.md} must have YAML frontmatter and a Markdown body:

\begin{Verbatim}[breaklines=true,breakanywhere=true,fontsize=\small]
---
name: lowercase-hyphenated-slug
description: What this skill does and when to trigger it. Include "NOT for: ..." exclusion conditions. 2-4 sentences.
category: general
---

<Markdown body with practical guidance>
\end{Verbatim}

\textbf{Step 5: Maintain Skill History}

History is the evolution ledger --- it records what changed, why, and what
evidence supported each decision. \textbf{Every action (create, improve,
optimize\_description) MUST leave a history trail.}

\textbf{CRITICAL: Read before write}

Before touching any existing skill, you \textbf{MUST}:

\begin{enumerate}
    \item Check whether \texttt{skills/<skill-name>/history/} exists; if it does, list it
    to see all existing entries.
    \item If it exists, read \textbf{every} \texttt{v*.md} and \texttt{v*\_evidence.md} file in that
    directory.
    \item If it exists, understand the full change trajectory before deciding your
    edit.
\end{enumerate}

Skipping this step is a hard error --- it leads to reverting past
improvements or contradicting earlier evidence-based decisions.

\textbf{History directory structure}

\begin{Verbatim}[breaklines=true,breakanywhere=true,fontsize=\small]
skills/<skill-name>/history/
├── v0_evidence.md ← why this skill was created (for create_skill)
├── v1.md          ← SKILL.md snapshot before round 1 edit
├── v1_evidence.md ← sessions/feedback that drove the v1→v2 change
├── v2.md          ← SKILL.md snapshot before round 2 edit
├── v2_evidence.md
└── ...
\end{Verbatim}

\textbf{History naming rules}

\begin{itemize}
    \item Use \textbf{version-based filenames only}: \texttt{v<N>.md} and \texttt{v<N>\_evidence.md}.
    \item \textbf{Do NOT} use dates, timestamps, or ad-hoc filenames such as
    \texttt{2026-04-04.md}, \texttt{notes.md}, or \texttt{new\_version.md}.
    \item Version numbers must reflect the evolution sequence of the skill, not the wall-clock date.
    \item If no history exists yet for an existing skill, the first snapshot you save is
    \texttt{v1.md} and the paired evidence file is \texttt{v1\_evidence.md}.
\end{itemize}

Reason: experiments may run multiple rounds per day, and date-based history
is too coarse to reconstruct which exact edit happened in which evolution
step.

\textbf{How to maintain history}

\textbf{For improve\_skill / optimize\_description:}
\begin{enumerate}
    \item Check \texttt{skills/<skill-name>/history/} to determine the current round N.
    If no history exists, this is round 1.
    \item Copy the current \texttt{SKILL.md} content verbatim to \texttt{history/v<N>.md}.
    \item Write \texttt{history/v<N>\_evidence.md} noting:
    \begin{itemize}
        \item Which sessions drove this change (session IDs, task IDs, PRM scores,
        success/fail counts, tool errors, repeated failure patterns)
        \item What the positive/negative signals were
        \item What previous history entries you read and how they informed this edit
        \item How the old version performed in the available session evidence
        \item Which exact sections/rules you are preserving, removing, or revising
        \item What action you decided (improve / optimize\_description)
    \end{itemize}
    \item Then edit \texttt{SKILL.md}.
\end{enumerate}

Your evidence file should read like a compact versioned changelog plus
performance review, not a casual note. Make it easy for a future agent to
answer:

\begin{itemize}
    \item Why did version \texttt{v<N>} need to change?
    \item What evidence from current sessions supports the next edit?
    \item How did prior versions appear to perform in historical sessions?
    \item Which modifications are intentional and should not be reverted casually?
\end{itemize}

\textbf{For create\_skill:}  
No previous version exists, but still write \texttt{history/v0\_evidence.md}
explaining:

\begin{itemize}
    \item What sessions motivated the creation (IDs, scores, failure patterns)
    \item Why no existing skill covers this pattern
    \item What action you decided (create\_skill)
\end{itemize}

\textbf{Evidence file content expectations}

Each \texttt{v<N>\_evidence.md} should include, in a concise but explicit form:

\begin{enumerate}
    \item \textbf{Decision summary}
    \begin{itemize}
        \item action type
        \item target skill
        \item why change is needed now
    \end{itemize}

    \item \textbf{Session evidence}
    \begin{itemize}
        \item relevant session IDs / task IDs
        \item representative PRM scores or aggregate metrics
        \item recurring tool failures / observations
    \end{itemize}

    \item \textbf{Historical comparison}
    \begin{itemize}
        \item what previous version(s) attempted
        \item whether later evidence suggests those edits improved outcomes,
        regressed outcomes, or remain inconclusive
    \end{itemize}

    \item \textbf{Edit plan}
    \begin{itemize}
        \item exact parts of the skill being changed
        \item exact parts intentionally preserved
    \end{itemize}

    \item \textbf{Open questions}
    \begin{itemize}
        \item uncertainty that future rounds should monitor
    \end{itemize}
\end{enumerate}

\textbf{History persistence depends on fresh mode}

\begin{itemize}
    \item In \texttt{--no-fresh} mode, the server refreshes \texttt{SKILL.md} from storage each
    round but does \textbf{NOT} clear the \texttt{history/} subdirectory. History therefore
    accumulates across rounds and serves as a continuous audit trail.
    \item In \texttt{--fresh} mode, the workspace is rebuilt from scratch each round, so
    local \texttt{history/} directories do \textbf{NOT} persist automatically. Treat each
    round as an isolated evolution pass unless the current workspace already
    contains history files.
\end{itemize}

\textbf{Editing Principles}

\textbf{Conservative Editing (for improve\_skill)}
\begin{itemize}
    \item Treat the CURRENT skill as the \textbf{source of truth}, not a rough draft.
    \item Default to \textbf{targeted edits}, not rewrites.
    \item Preserve the original structure, heading order, and terminology.
    \item If failures are only corner cases, add missing checks or clarify
    constraints without changing unrelated sections.
    \item Only rewrite an entire section if evidence shows it is materially wrong.
    \item If a successful session supports a section, leave it untouched unless
    failure evidence explicitly contradicts it.
\end{itemize}

\textbf{Hard Constraints}
\begin{itemize}
    \item Do \textbf{NOT} change API contracts, ports, endpoints, output paths, payload
    formats, or required filenames --- unless session evidence clearly shows
    they have changed.
    \item Do \textbf{NOT} remove core capabilities, API references, or tool-usage examples
    unrelated to observed failures.
    \item Do \textbf{NOT} turn a skill into a different skill with a different purpose.
    \item Do \textbf{NOT} rewrite the whole skill from scratch.
    \item Do \textbf{NOT} impose a new template or writing style unless evidence requires it.
    \item Do \textbf{NOT} add generic best-practice guidance (retry logic, caching, state
    management) unless the environment has specific quirks.
\end{itemize}

\textbf{Distinguishing Skill vs Agent Problems}

Not every failure is a skill deficiency:

\begin{itemize}
    \item \textbf{Skill problem} (wrong/missing guidance) $\rightarrow$ edit the skill.
    \item \textbf{Agent problem} (misuse, restarts, context overflow) $\rightarrow$ do \textbf{NOT} bloat the
    skill with agent-runtime advice.
    \item \textbf{Environment problem} (API instability, network flakiness) $\rightarrow$ add a brief
    note if recurrent, but keep it short.
\end{itemize}

Critical anti-pattern: if the skill \textbf{ALREADY} contains correct environment
information and the agent failed because it did \textbf{NOT} use that information,
that is an \textbf{AGENT} problem. Do \textbf{NOT} delete correct API info and replace it with
instructions like ``go inspect the source code''.

\textbf{Skill Writing Principles (for create\_skill)}

\begin{itemize}
    \item A skill should compress \textbf{environment information} (API endpoints, ports,
    payload formats, tool quirks, domain procedures) --- not generic best
    practices the agent already knows.
    \item Prefer a short, action-oriented name (lowercase-hyphenated slug).
    \item The name \textbf{MUST} differ from all existing skills. Check \texttt{manifest.json} for
    the current list of skill names before creating a new one.
    \item Description is the main triggering mechanism --- put clear triggering
    contexts there, including ``NOT for: ...'' exclusion conditions.
    \item Content should be domain-specific, practically useful, and non-obvious.
    \item Use imperative instructions. Organize the body naturally for the task.
    \item Include concrete API endpoints, ports, command patterns, and payload
    examples when they are central to the task.
    \item Keep it concise, reusable, and evidence-driven.
    \item Write reusable guidance, not a failure summary or postmortem.
\end{itemize}

\textbf{Important Notes}

\begin{itemize}
    \item You may create multiple skills in one session if the evidence supports it.
    \item Process \textbf{ALL} sessions --- don't stop after the first group.
    \item Write your changes directly to files in \texttt{skills/}. The server will detect
    what changed by comparing file hashes.
    \item \textbf{ALWAYS} read \textbf{ALL} files in \texttt{skills/<name>/history/} before deciding any
    action on that skill, if that history directory exists. This is
    mandatory, not optional.
    \item \textbf{ALWAYS} save the old version and evidence before making changes.
    \item \textbf{ALWAYS} use version-based history filenames (\texttt{v<N>.md},
    \texttt{v<N>\_evidence.md}); never use date-based filenames.
    \item Do \textbf{NOT} modify files in \texttt{sessions/} --- they are read-only input.
    \item Do \textbf{NOT} modify \texttt{manifest.json} or \texttt{skill\_registry.json} --- the server
    manages those.
    \item Do \textbf{NOT} access files outside this workspace directory.
    \item If there are no actionable patterns in the sessions, it is perfectly fine
    to make no changes at all.
\end{itemize}
\end{templatebox}